\documentclass{article}

\PassOptionsToPackage{numbers, compress}{natbib}

\usepackage[preprint]{nips_2018}
\usepackage{floatrow}  
\usepackage[utf8]{inputenc} 
\usepackage[T1]{fontenc}    
\usepackage{hyperref}       
\usepackage{url}            
\usepackage{booktabs}       
\usepackage{amsfonts}       
\usepackage{nicefrac}       
\usepackage{microtype}      
\usepackage{graphicx} 
\usepackage{amsmath}
\usepackage{xcolor}
\usepackage{amssymb}
\usepackage{hyperref}
\usepackage{wrapfig}
\usepackage{natbib}

\makeatletter
\g@addto@macro\normalsize{%
  \setlength\abovedisplayskip{3pt}
  \setlength\belowdisplayskip{3pt}
  \setlength\abovedisplayshortskip{3pt}
  \setlength\belowdisplayshortskip{3pt}
}

\usepackage[bbgreekl]{mathbbol}

\usepackage{lipsum}
\usepackage{titlesec}

\titlespacing\section{0pt}{-1pt plus 4pt minus 2pt}{0pt plus 2pt minus 2pt}
\titlespacing\subsection{0pt}{-1pt plus 4pt minus 2pt}{0pt plus 2pt minus 2pt}
\titlespacing\subsubsection{0pt}{-1pt plus 4pt minus 2pt}{0pt plus 2pt minus 2pt}

\title{Deep Graph Translation}
 \author{
   Xiaojie Guo\\
   George Mason University\\
   Fairfax, VA 22030 \\
   \texttt{xguo7@gmu.edu} \\
   \And
   Lingfei Wu \\
   IBM Research\\
   Yorktown Heights, NY 10598\\
   \texttt{wuli@us.ibm.com} \\
   \And
   Liang Zhao \\
   George Mason University\\
   Fairfax, VA 22030 \\
   \texttt{lzhao9@gmu.edu} \\
 }

\begin{document}

\maketitle
\vspace{-0.6cm}
\begin{abstract}
Inspired by the tremendous success of deep generative models on generating continuous data like image and audio, in the most recent year, few deep graph generative models have been proposed to generate discrete data such as graphs. They are typically unconditioned generative models which has no control on modes of the graphs being generated. Differently, in this paper, we are interested in a new problem named \emph{Deep Graph Translation}: given an input graph, we want to infer a target graph based on their underlying (both global and local) translation mapping. Graph translation could be highly desirable in many applications such as disaster management and rare event forecasting, where the rare and abnormal graph patterns (e.g., traffic congestions and terrorism events) will be inferred prior to their occurrence even without historical data on the abnormal patterns for this graph (e.g., a road network or human contact network). To achieve this, we propose a novel Graph-Translation-Generative Adversarial Networks (GT-GAN) which will generate a graph translator from input to target graphs. GT-GAN consists of a graph translator where we propose new graph convolution and deconvolution layers to learn the global and local translation mapping. A new conditional graph discriminator has also been proposed to classify target graphs by conditioning on input graphs. Extensive experiments on multiple synthetic and real-world datasets demonstrate the effectiveness and scalability of the proposed GT-GAN.
\end{abstract}

\section{Introduction}

Graphs are universal representations of pairwise information in many problem domains such as social networks \cite{wang2017community}, questions answering \cite{bordes2014question}, molecule design \cite{kusner2017grammar}, and quantum chemistry \cite{gilmer2017neural}. To improve the performance of graph-based learning tasks, many recent research efforts have concentrated on encoding a graph (or graph nodes) into a vector representation, either through solving graph embedding tasks \cite{perozzi2014deepwalk,tang2015line,grover2016node2vec,hamilton2017inductive} or developing graph neural networks by extending well-known network architectures (e.g., convolutional and recurrent neural networks) to graph data \cite{scarselli2009graph,li2016gated,bruna2014spectral,defferrard2016convolutional,kipf2017semi}.

Very recently there has been a surge of interests in deep generative models that decode from a vector representation to discrete data such as graphs, molecules, and computer programs \cite{gomez2016automatic,kusner2017grammar,dai2018syntax,simonovsky2018graphvae,li2018learning,samanta2018designing,jin2018junction}, in part because of the recent successes in learning coherent latent representations for continuous data of image, video, and audio. There are several unique challenges in graph generations: i) using strings to represent graphs like molecules is brittle due to its sensitivity of small changes in the string leading to complete different molecules; ii) learning order for incremental construction of graph nodes and edges is difficult; iii) graphs are discrete structure with variable sizes and permutation invariant. To (partially) address these challenges, recently proposed approaches are either focusing on incorporating the structural constraints like context-free grammars into the generative models \cite{kusner2017grammar,dai2018syntax} or taking advantage of recently developed graph neural networks to directly express on graphs rather than linear string representations \cite{simonovsky2018graphvae,li2018learning,samanta2018designing,jin2018junction}. 

In the most recent year, although there are promising progresses in graph generations, the existing approaches are typically unconditioned generative models, which has no control on modes of the graphs being generated. However, it is usually desirable to guide the graph generation process by conditioning the model on additional information such as data from the different modality. In many real-world applications, we are interested in generating a new graph conditioning on an input graph. Examples include generating the traffic jam situation given a road network without traffic jam, and inferring the contact network when specific type of emergency occurs (e.g., terrorism events) given a people-people contacting network in normal situation, especially the anomaly event has not appeared before on the specific input graph. Our motivating application comes from the enterprise-network user authorization which requires predicting the hacked authentication network for the users who have not been hacked before. We formulate this type of new problems as \emph{Deep Graph Translation}, which aims at translating a graph with one modality to a new one with other modality using deep neural networks architecture. This problem is analogical to image-to-image translation in image processing \cite{isola2017image} and language translation in natural language processing \cite{yang2017improving}. Unfortunately, existing graph generation approaches and image or text translation methods cannot be directly applied to the NGT due to the aforementioned challenges.  

In this paper, we present a novel framework for deep graph translation with graph-translation generative adversarial nets (GT-GAN). Different from existing ordinary GANs \cite{goodfellow2014generative} that learn generative model of data distribution, GT-GANs learn a conditional generative model, which is a graph translator that is conditioned on input graph and generate the associated target graph. The GT-GAN consists of a graph translator and a conditional graph discriminator. The graph translator includes two parts - graph encoder and graph decoder, where we extend a recent proposed graph convolutional neural nets \cite{kawahara2017brainnetcnn} for the graph encoder and propose a new graph decoder to address not only the previous mentioned challenges for graph generation but also new challenges brought by graph translation on graph local property preservation. We also propose a new conditional graph discriminator that is built on graph convolutional neural nets (GCNN) but conditioning on the input graphs. In addition, Our proposed GT-GAN is highly scalable whose time and memory complexity are equal or better than quadratic with the number of nodes, which makes it suitable for at least modest-scale graph translation problems. More importantly, our GT-GAN is highly extensible where underlying building blocks, GCNN and distance measure in discriminator, can be replaced by any advance techniques such as \cite{kipf2017semi,arjovsky2017wasserstein}. Extensive experiments on both synthetic and real-world application data demonstrate that GT-GAN is capable of generating graphs close to ground-truth target graphs and significantly outperforms other VAE-based models in terms of both effectiveness and efficiency.

\section{Related Works}

Modern deep learning techniques for research on graphs is a new trending topic in recent years.



\textbf{Neural networks on graph representation learning}. In recent years, there has been a surge of research in neural networks on graphs, which can be generally divided into two categories: Graph Recurrent Networks \cite{gori2005new,scarselli2009graph,li2016gated} and Graph Convolutional Networks \cite{niepert2016learning,mousavi2017hierarchical,defferrard2016convolutional,kawahara2017brainnetcnn,nikolentzos2017kernel,cao2016deep,kipf2017semi,xu2018graph2seq}. Graph Recurrent Networks originates from early works for graph neural networks that were proposed by Gori et al. \cite{gori2005new} and Scarselli et al. \cite{scarselli2009graph} based on recursive neural networks. Graph neural networks were then extended using modern deep leaning techniques such as gated recurrent units \cite{li2016gated}. Another line of research is to generalize convolutional neural networks from grids (e.g., images) to generic graphs. Bruna et al. \cite{bruna2014spectral} first introduces the spectral grpah convolutional neural networks, and then extended by Defferrard et al. \cite{defferrard2016convolutional} using fast localized convolutions, which is further approximated for an efficient architecture for a semi-supervised setting \cite{kipf2017semi}. For most of them, the input signal is given over node with static set of edge and their weights fixed for all samples. To consider the graph topology, our graph encoder draws inspiration from \cite{kawahara2017brainnetcnn} that presents a new CNN consisting of a sequence of new layers for the convolution operations on edge-to-edge, edge-to-node, and node-to-graph for undirected graphs. 



\textbf{Graph generation}. Deep model based graph generation has attracted fast increasing attentions very recently, and is still largely unexplored. Until now there are paucity of deep models proposed for graph generation, which can be generally categorized into two classes: 1) domain-specific models, and 2) generic models. Domain-specific models \cite{gomez2016automatic,kusner2017grammar,dai2018syntax,jin2018junction} typically take sequence inputs (such as SMILES for representing molecules) and generate another sequences or parse trees from context-free grammars or sub-graphs by utilizing the collection of valid components, by leveraging the celebrated Seq2Seq model \cite{sutskever2014sequence} based on variational autoencoder (VAE) \cite{kingma2013auto}. However, these methods are highly tailored to only address the graph generation in specific type of applications such as molecules generation.
Generic graph generation can handle general graphs that is not restricted into specific applications, which is more relevant to this paper. Existing works are basically all proposed in the most recent year, which are based on VAE \cite{simonovsky2018graphvae,samanta2018designing}, generative adversarial nets (GAN) \cite{bojchevski2018netgan} and others \cite{li2018learning}. Specifically, Li et al. \cite{li2018learning} proposed a graph net that generates nodes and edges sequentially to form a whole graph, which is sensitive to the generation order and time consuming for large graph. Similarly, Bojchevski et al. \cite{bojchevski2018netgan} also prefer to generate nodes and edges sequentially, by random walk on the graphs which also faces similar problems. Different from the above methods, Simonovsky et al. \cite{simonovsky2018graphvae} and Samanta et al. \cite{samanta2018designing} both propose new variational autoencoders in parallel for whole graph generation, though they typically only handle very small graphs (e.g., with $\le 50$ nodes) and cannot scale well in both memory and runtime for large graphs.

\textbf{Conditional generative adversarial nets}. Generative adversarial networks (GANs) \cite{goodfellow2014generative} significantly advanced the state
of the art over the classic prescribed approaches like mixtures
of Gaussians for image and 3D object generation \cite{karras2017progressive}. To further empower GANs with the capability to generate objects under certain conditions, recently conditional GANs have been proposed to condition GANs on discrete labels \cite{mirza2014conditional}, text \cite{reed2016generative}, and especially images \cite{wang2016generative}. Conditional GANs enable multi-modal data fusion to generate images according to specific annotations, text labels, or translate an input image to an output image. For example, many important applications in image processing domain using conditional GANs include image-to-image translation \cite{isola2017image}, inpainting \cite{pathak2016context}, and style transfer \cite{zhu2017unpaired}. However, currently, to the best of our knowledge, there is little to no work focusing on graph-to-graph translation problem.

\section{GT-GAN}
We first present the problem formulation of graph translation. And then propose our method GT-GAN and describe its graph translator and conditional graph discriminator in details.

\subsection{Deep Graph Translation by GT-GAN}
This paper focuses on graph translation from an \emph{input graph} to a \emph{target graph}. We define an \emph{input graph} $X$: $G_X=(V,E,A)$ as a directed weighted graph such that $V$ is the set of $N$ nodes, $E\subseteq V\times V$ is the set of directed edges, and $A\in\mathbb{R}^{n\times n}$ is a matrix holding the set of weights for the corresponding edges, called \emph{weighted adjacency matrix}. Denote $e_{i,j}\in E$ as an edge from the node $v_i\in V$ to  $v_j\in V$ and thus $w_{i,j}\in W$ is used to denote the corresponding weight of the edge $e_{i,j}$. Similarly, we define a \emph{target graph} $G_Y=(V',E',A')$ also as directed weighted graph.
Typically we focus on learning the translation from one topological patterns to the other one the same set of nodes and hence we have $V=V'$. Additionally, the weights of edges are nonnegative such that $A_{i,j}\ge 0$ and $A_{i,j}'\ge 0$. 
We define a new problem named \emph{\textbf{graph translation}}, where we focus on learning a translator from an input graph $G_X\in\mathcal{G}_X$ and a random noise $U$ to a target graph $G_Y\in\mathcal{G}_Y$, the translation mapping is denoted as $T:U,G_X\rightarrow G_Y$, where $\mathcal{G}_X$ and $\mathcal{G}_Y$ denote the domains of input and and target graphs, respectively.

\begin{figure}
  \setlength{\abovecaptionskip}{0pt}
  \setlength{\belowcaptionskip}{0pt}
  \centering\vspace{-0.35cm}
  \captionsetup{justification=centering}
  \includegraphics[width=0.95\textwidth] {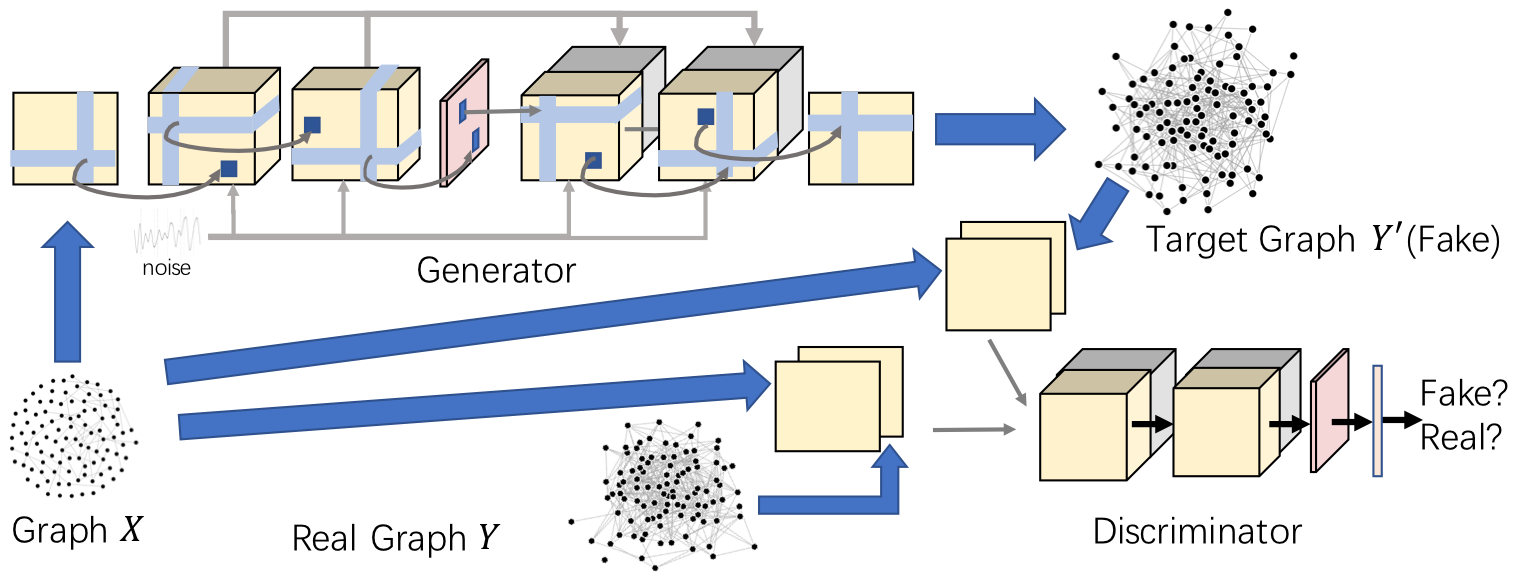}
  \vspace{-0.3cm}\caption{Architecture of GT-GAN consisting of graph translator and conditional graph discriminator. New graph encoder and decoder are specially designed for graph translation.\vspace{-0.4cm}}
  \label{fig:GT-GAN}
\end{figure}

To address this issue, we proposed the Graph-Translation GAN (GT-GAN) that consists of graph translator $T$ and conditional graph discriminator $D$, as shown in Figure \ref{fig:GT-GAN}. $T$ is trained to produce target graphs that cannot be distringuished from ``real'' images by $D$, distinguish the produced target graph $G_{Y'}=T(G_X,U)$ from the real one $G_Y$ based on the current input graph $G_X$. $T$ and $D$ undergo an adversarial training process based on input and target graphs shown as below:
\begin{align}
    \mathcal{L}(T,D)&=\mathbb{E}_{G_X,G_Y}[\log D(G_Y|G_X)]+\mathbb{E}_{G_X,U}[\log(1-D(T(G_X,U)|G_X))]
\end{align}
where $T$ tries to minimize this objective against an adversarial
$D$ that tries to maximize it, i.e. $T^* =
\arg\min_T \max_D \mathcal{L}(T,D)$.

\subsection{Graph Translator}

We propose a new graph translator by extending the state-of-the-art graph convolution and proposing new graph deconvolution layers.




Different and more difficult than graph generation, for graph translation, not only the latent graph representation but also the mapping from input graph to the target graph need to be learned. However, graph representations vary according to different graph samples while the mapping between input-target pairs remains consistent across different pairs of samples. Thus the ordinary encoder-decoder architecture is not sufficient because it is difficult to jointly compress both the sample-specific representation and sample-invariant mapping into a unified embedding layer with good model generalization. Moreover,  different from graph generation which focus on the graph level distribution, graph translation requires to learn the translation mapping both in graph level (e.g., general change in graph metrics) and in local level (e.g., specific nodes have specific changes according to its local properties).

To address the first challenge, we propose to leverage skip-net structure \cite{ronneberger2015u} so that the sample-specific representations can be directly passed over through skip connection to decoder's layers while the sample invariant mapping will be learned in the encoder-decoder structure. More importantly, to address the second challenge, we extend the state-of-the-art graph convolution and propose new graph deconvolution that preserve both global and local information during graph translation. The architecture of graph translator is illustrated in Figure \ref{fig:GT-GAN}, where the input graph first undergoes two ``directed edge-to-edge convolution' operations to encode the higher-order topological information and then is embedded into node representation by a ``directed edge-to-node convolution'' operation. Then, a novel graph decoder based on graph deconvolution has been proposed, including one ``directed node-to-edge deconvolution'' operation and two ``directed edge-to-edge deconvolution'' operations.  The details are introduced in the following.


\subsubsection{Directed Graph Convolutions}






Some recent works focus on generalizing image convolution into graph convolution, but as pointed out in \cite{kawahara2017brainnetcnn}, in most of them the input signal is given over node with static set of edge and their weights fixed for all samples. The work that can handle encode the edge structure typically focus on undirected graphs but are not immediately applicable to directed graphs. 

Therefore, to solve it, we propose our method on directed graph convolution. In a directed weighted graph, each node could have in-edge(s) and out-edge(s), which can be convoluted respectively for each node. Specifically, denote the $Z^{l,m}\in\mathbb{R}^{N\times N}$ as the weighted adjacency matrix for $l$th layer in $m$th feature map and $Z_{i,j}^{l,m}$ is for the edge $e_{i,j}$. Let $Z^{1,1}\equiv A$ denote the weighted adjacency matrix of the input graph. $\Phi_{l,m}\in\mathbb{R}^{1\times N}$ and $\Psi_{l,m}\in\mathbb{R}^{N\times 1}$ are the incoming and outgoing kernels of $l$th layer of $k$ feature map for a node, respectively. We define a graph convolution over its in-edge(s) as the weighted sum over all the weights of its incoming edges: $f^{\mbox{\scriptsize (in)}}_{l,m,n,j}=\Phi_{l,m,n}\cdot A_{\cdot,j}^{l,m}$. Similarly, we define the graph convolution over the out-edge(s) as $f_{l,m,n,i}^{\mbox{\scriptsize (out)}}=A^{l,m}_{i,\cdot}\cdot\Psi_{l,m,n}$. And thus the \emph{{directed edge-to-edge convolution}} is defined as follows: 
\vspace{-0.15cm}\begin{equation}
A^{l+1,n}_{i,j}=\sigma(\sum\nolimits_{m=1}^{M_l}(f_{l,m,n,i}^{\mbox{\scriptsize (in)}}+f_{l,m,n,j}^{\mbox{\scriptsize (out)}}))
\end{equation}
where $A_{i,j}^{l+1,m}$ refers to the $m$th  value in position ${i,j}$ of edge level feature map in the $(l+1)$th layer. $M_l$ refers to the number of feature maps in the $l$th layer. The two components of the formula refers to direction filters as talked above. $\sigma(\cdot)$ refers to activation function that can be set as linear, or ReLu \cite{velivckovic2017graph} when the edge weights are assumed nonnegative.

The \emph{directed edge-to-node convolution} embeds each edge feature map into a vector which encodes all the incoming and outgoing edges of a node into a value from various combinations:
\begin{align}
    A_i^{4,n}=\sigma(\sum\nolimits_{m=1}^{M_3} (f_{3,m,n,i}^{\mbox{\scriptsize (in)}}+f_{3,m,n,i}^{\mbox{\scriptsize (out)}})) 
\end{align}
where $A_i^{4,n}\in\mathbb{R}^{n\times 1}$ denotes the $i$th node representation (i.e., the $4$th layer in graph translator in Figure \ref{fig:GT-GAN}) under $n$th feature map.





\subsubsection{Graph Deconvolutions}
\label{sec:graph_decoder}

\begin{wrapfigure}{r}{0.4\textwidth}
  \includegraphics[width=\textwidth] {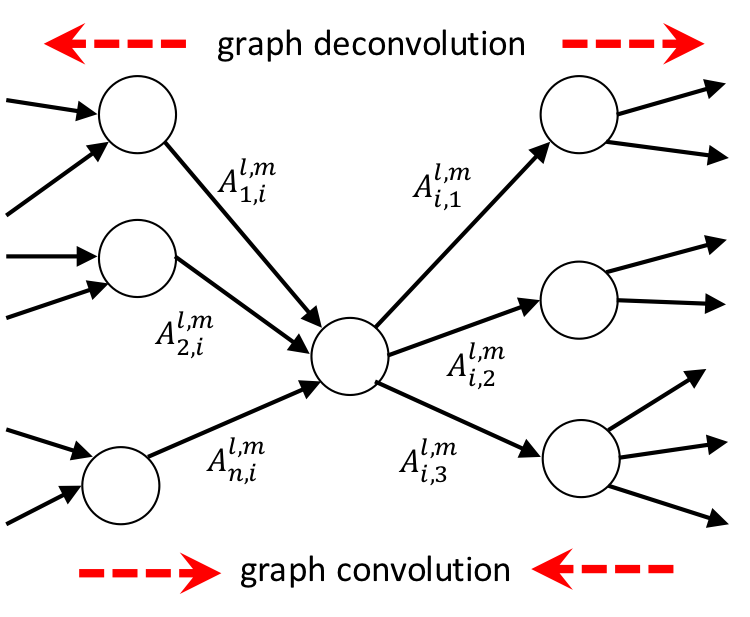}
  \captionsetup{justification=centering}
  \caption{ Graph deconvolution decodes the single node (or edge) information to its incoming and outgoing neighbor nodes and edges, which is a reverse process of graph convolution.}
  \label{fig:layers}
\end{wrapfigure}

The graph convolution encodes the input graph into node representation with highly condensed knowledge on the higher-order neighborhood pattern of node. Next, graph translator requires to deconvolute the node representation back to target graph. Currently, to the best of our knowledge, there is little to no work on directed graph deconvolution. The techniques that can generalize the image deconvolution to graph domain is
highly imperative and nontrivial. This paper proposes novel graph deconvolution techniques including ``node-to-edge deconvolution'' and ``edge-to-edge deconvolution''. 
Different from images, ``directed node-to-edge deconvolution'' requires to decode the ``local neighborhood'' pattern from a node through graph topology including both incoming and outgoing connections. This calls for a reverse process of
``directed node-to-edge convolution'' as shown in Figure \ref{fig:layers}. To achieve this, the node representation  $A^{l,k}\in\mathbb{R}^{1\times n}$ in $l$th layer in $k$th feature map will be multiplied by the transpose of incoming and outgoing kernels to obtain weighted adjacency matrix in the $(l+1)$th layer:
\vspace{-0.25cm}\small
\begin{align}
\label{eq:n2edeconv}
        A^{l+1,m}_{i,j}=\sum\nolimits^{M}_{k=1}\sigma([\Phi_{l,k}^T\cdot A_j^{l,k}]_i+[ A_{i}^{l,k}]_j\cdot \Psi_{l,k}^T)
\end{align}\normalsize where $[\cdot]_i$ means the $i$th element of a vector. 

The decoded edges from node representation still encompasses highly-ordered connectivity knowledge, which will be released and translated back to the neighborhood of incident incoming and outgoing edges by ``edge-to-edge deconvolution'':\small
\begin{equation}
\label{eq:e2edeconv}
    A^{l+1,m}_{i,j}=\sigma([\sum\nolimits_{k=1}^n\Phi_{l,m}^T\cdot A_{k,j}^{l,m}]_i+[\sum\nolimits_{k=1}^n A_{i,k}^{l,m}\cdot \Psi_{l,m}^T]_j)\ \ \ \ \ \ \ \ \ \ \ \ 
\end{equation}\normalsize
Actually, graph deconvolution can also be seen as a transpose operation of graph convolution, as shown in Equations \eqref{eq:e2edeconv} and Equations \eqref{eq:n2edeconv}, which is analogical to the relationship between image convolution and deconvolution\footnote{This is why ``deconvolution'' is well-recognized to be better named as ``transposed convolution''}.

\subsection{Conditional Graph Discriminator}
When identifying the ``translated'' target graph from the ``real'' ones, the input graph will also need to be considered. This new problem requires the discriminator to take two graphs as inputs, namely a target graph and a input graph, and classify the joint patterns of them into a class label. Instead of identifying whether the generated graphs look similar to the real graphs, this problem require to identify whether the translations look similar, which encodes a second-order change between input and target graphs. The challenge is that we need not only to learn the topological representation of each graph but also need to learn the correspondence between the two graphs and their nodes globally and locally. To address this, we propose \emph{conditional graph discriminator} (CGD) which leverages ``paired'' directed edge-to-edge layers as shown in Figure \ref{fig:GT-GAN}. Specifically, the input and target graphs are both ingested by CGD and stacked into a $N\times N\times 2$ tensor which can be considered as a 2-channel weighted adjacency matrix of a multi-graph. Then each of the channels are mapped to its corresponding feature maps and then the separated directed edge-to-edge layers. The the edge-to-node layers is again applied to obtain node representations, which is then mapped to graph embedding by a fully-connected layer. Finally, a softmax layer is implemented to distinguish the real graph and fake graph. Both the generator and discriminator are trained through ADAM optimization algorithm.




\subsection{Analysis on Time and Memory complexity}
The graph encoder and decoder shares the same time complexity. Without loss of generalization, we assume all the layers (except input and output) have the same number of feature maps as $M_0$. $P$ is the length of the fully connected layer. Then, the worst-case (i.e., when the weighted adjacency matrix is dense) total complexity of GT-GAN is $O(9 N^{2}M_0^2+3N^2 M_0^2+N^2 M_0 P)$, where the first, second, and third terms are for all the ``directed edge-to-edge convolutions'', ``directed edge-to-node convolutions'', and fully connected layers in conditional graph discriminator, respectively.
The total memory complexity of GT-GAN is  $O((9NM^{2}_{0}+9N^{2}M_{0}) +(3NM_{0}^{2}+3NM_{0}) +(N^{2}M_{0}P+P))$, for all the ``directed edge-to-edge convolutions'', ``directed edge-to-node convolutions'', and fully connected layers in conditional graph discriminator, respectively. 


\section{Experiment}
In this section, extensive experiments on graph translation for the proposed methods on two synthetic and one real-world datasets are conducted and analyzed. The proposed GT-GAN demonstrated outstanding performance in both effectiveness and scalability by multiple metrics. All experiments are conducted on a 64-bit machine with Navida GPU (GTX 1070,1683 MHz, 8 GB GDDR5).

\subsection{Experimental Settings}
\subsubsection{Datasets}
\label{sec:datasets}
Two groups of synthetic datasets are used to validate the performance of the proposed GT-GAN. The first group is based on scale-free graphs while the second is based on Poisson-random graphs. Each group has three subsets with different graphs size (number of nodes): 50,100 and 150. Each subset consists of 5000 pairs of input and target graphs: 2000 pairs of graphs are used for training the graph translator and the remaining 3000 pairs are used as test set to validate the 3000 pairs generated by the trained translator. The generation rule for each group of datasets is as follows.

\textbf{Scale-free Graph Datasets} Here each input graph is generated as a scale-free network, which is a network whose degree distribution follows a power law \cite{bollobas2003directed}. Here, 0.54 is set as the probability for adding an edge between two existing nodes, where one is chosen randomly according the in-degree distribution and the other according to the out-degree distribution. Thus, both input and target graph are scale-free graphs.

\textbf{Poisson-random Graph Datasets}. The input graph is generated by Barabasi-Albert model \cite{albert2002statistical} which is another kind of scale-free network. Then for an input graph with number of $|E|$ edges, we randomly add another $k\cdot |E|$ edges on it to form the target graph, where $k$ follows the Poisson distribution with mean of $5$.

\textbf{Real World Datasets: User Authorization Datasets}.
This dataset includes the authentication activities of users on their computers and servers in an enterprise computer network and has been publicly released by LANL \cite{kent-2015-cyberdata1}. In a period of time, each user logs on different computers from other computers using credential authentication, which forms a directed weighted graph: nodes represent computers that are available to a specific user while the directed edges and weights represent the authentication actions and the frequencies from one computer to another. Such user authentication graph can reflect whether the user's account has been hacked or not since the hacker's authentication behavior could be highly different than this original user. 
Our datasets consist of positive and negative classes, namely hacked (i.e, malicious) graphs and normal graphs. Each hacked graph contains both the activities from the hacker(s) and the user while its corresponding normal graph only contains the activities of the user. By learning the graph translation between normal and hack graphs, we learn the hackers' generic behaviors across different graphs.
There are 97 users, and each user has its own samples of input graphs and target graphs. There are two datasets: one has 315 pairs of graphs with 300 nodes and another has 315 pairs of graphs with 50 nodes. Graphs in each subset are split into 3 parts for cross validation. Two parts are used to train a translator and this translator is used to generate the malicious graphs for the third part. More details about this real dataset can be found in Appendix C.

\subsubsection{Evaluation Metrics}
\paragraph{Direct Evaluation}
\label{p:direct}
To directly verify if GT-GAN indeed can discover the underlying ground-truth translation rules between input and target graphs, we examined the degree distribution and the ratio of added edges for scale-free and poisson-random graph datasets, respectively. Specifically, for scale-free graph datasets, the metrics used to measure the distance between the real degree distributions and generated graph degree distribution are: Jensen-Shannon distances (JS) \cite{lin1991divergence}, Hellinger Distance (HD) \cite{beran1977minimum}, Bhattacharyya Distance (BD) \cite{basseville1989distance} and Wasserstein Distances (WD) \cite{ruschendorf1985wasserstein}. For Poisson-random graph, the distributions of $k$ in the real target graphs and those generated graphs are compared.

For the user authorization dataset, the real target graphs and those generated by GT-GAN and comparison method are compared under well-recognized graph metrics including degree of nodes, reciprocity, and density.  We calculate the distance of degree distribution and Mean Sqaured Error (MSE) for reciprocity and density.
\paragraph{Indirect Evaluation.}
\label{p:indirect}
One important goal of graph translation is to generate the target graph for any new-coming input graph. Take our real-world dataset as an instance, user authentication network can be used to classify whether a user's account is malicious or not by examining her historical malicious- and normal-situations of authentication network. However, this classification cannot be achieved for those users which have no malicious situation before. This can be handled by graph translator which learns the translation mapping based on all the other users and then simulate the malicious situation for those without them before. 
Thus, we use the generated graphs to train a classifier, and if this classifier can classify accurately on real input and target graphs, it  demonstrates its effectiveness. Specifically, first, train the graph translator based on training set and generate the graphs for input graphs in test set. Then, split the test set into two parts. First part is used to train two graph classifiers that distinguish input graphs from target graphs. The first classifier is trained by generated target graphs while the second is trained by real target graphs. Then the second part test data is used to test and compare the classification performance of these two classifiers. If the first classifier's performance is good and near to the second one, it demonstrates the effectiveness of the graph translator.

\begin{table}[h]\scriptsize
  \centering
  \label{table:node degree}
  \caption{Distance of node degree distribution between generated graphs and real graphs}
  \begin{tabular}{llllll}\\\hline\hline
    Graph size &Methods &Jensen-Shannon &Hellinger &Bhattacharyya &Wasserstein \\\hline\hline
    50& R-VAE & 0.69&0.99&Inf&7.99 \\
    50 &GT-GAN &0.43&0.89&1.66&2.43\\
    \hline
    100&GT-GAN&0.15&0.43&0.24&0.31\\
    \hline
    150& GT-GAN &0.07&0.30&0.11&0.27\\\hline\hline
  \end{tabular}  
\end{table}
\subsubsection{Comparison Methods}

There is no existing work on deep graph translation. Prior to it, deep graph generation attracts big interests especially in the most recent half year. There are few existing methods for generic graph generation \cite{samanta2018designing, simonovsky2018graphvae, li2018learning, dai2018syntax} which learns to mimic a similar graph by learning a distribution of historical graphs. Most of the methods were proposed in recent quarter of year and only the code of randomVAE  \cite{samanta2018designing} is publicly available for comparison experiment. RandomVAE is a general graph generation method based on variational auto-encoder. R-VAE generates each node sequentially conditioning on the previously generated nodes. Here, since R-VAE cannot condition on input graphs, it directly learns the distribution of real target graphs and generate new ones. 

\subsection{Performance}
\subsubsection{Model scalability Analysis}
\label{paragraph:scalability}
\begin{figure}
  \centering\vspace{-0.3cm}
  \includegraphics[width=0.9\textwidth]{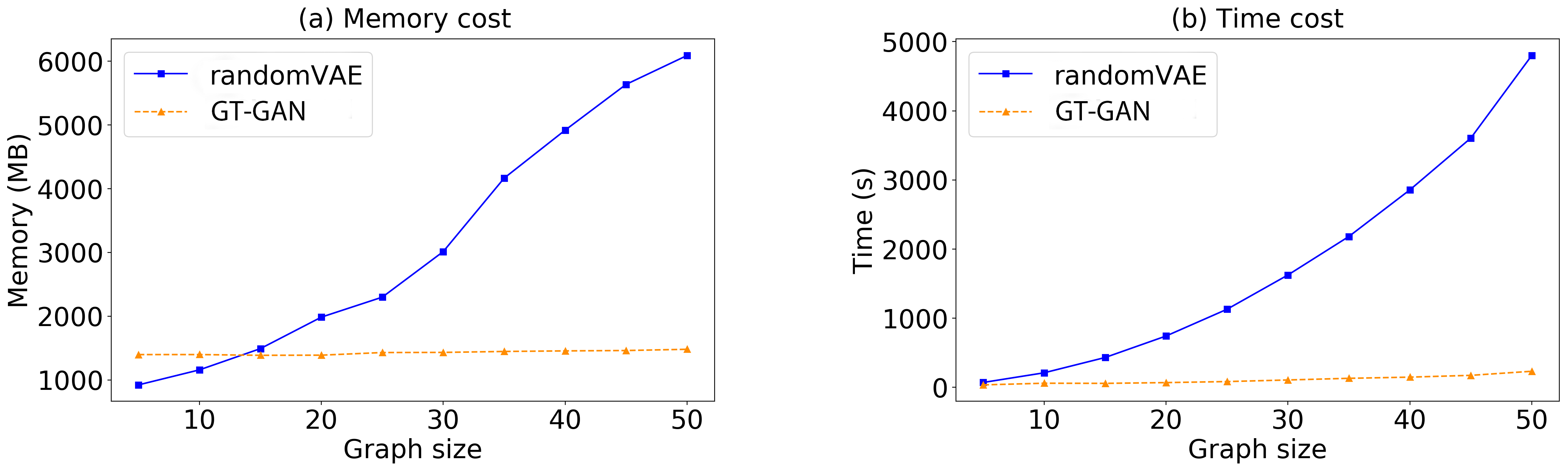}
  \caption{\vspace{-0.4cm} Scalability plot on memory and time}
  \label{salability}
\end{figure}

The proposed GT-GAN has the ability to handle graphs with much larger size compared to randomVAE. To demonstrate it, Figure\ref{salability} illustrates the scalability of GT-GAN and randomVAE in terms of memory and time cost, respectively. The memory cost of our GT-GAN is very stable at about 1500MB when the graph size increase from 1 to 50, while the memory cost of randomVAE increases by 5000MB. The training time of GT-GAN increases slightly due to the mini-batch training algorithm, while the time cost of randomVAE increases quickly by 5000s from size 1 to size 50 because it only learns one graph at each iteration.

\subsubsection{Performance on Scale-Free Graph Datasets}
\paragraph{Direct Evaluation}
Graphs are generated by GT-GAN and randomVAE on test set. The node degree distributions of the generated graphs are compared with those of real target graphs in several distances metrics Table \ref{table:node degree}. Graph Size means the number of nodes. Graphs tested on GT-GAN have size from 50 to 300, while randomVAE is only tested on graph size of 50 due to its prohibitive costs on the memory and time even in modest size of graphs (e.g., size of 100). \ref{paragraph:scalability}. 


The proposed GT-GAN significantly outperforms randomVAE in all metrics, especially in Bhattacharyya distance and Wasserstein distance. The ``Inf'' in Tabel \ref{table:node degree} represents the distance more than 1000. The Bhattacharyya distance of randomVAE is large because randomVAE cannot tailor the generated target graph according to its corresponding input graph, which will not work well when the distribution of target graph changes largely. The performance of GT-GAN becomes even better as the graph size increases because the statistical properties of the generated graphs become more obvious in larger graphs.  

\begin{figure}
  \centering
  \label{figure:synthetic2}
    \includegraphics[width=0.9\textwidth]{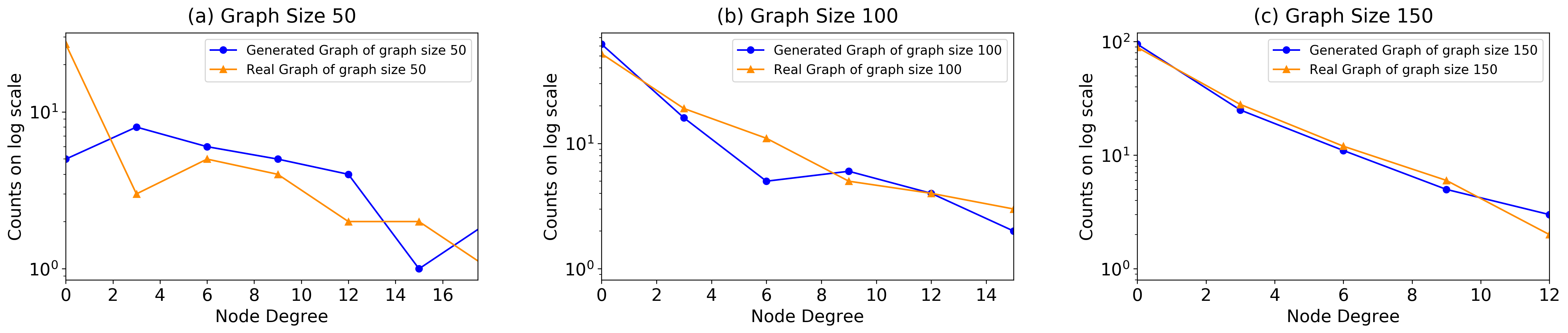}
\caption{Examples of node degree distributions of generated and target graphs in different sizes }
\end{figure}

Figure \ref{figure:synthetic2} directly shows the node degree distribution curve of some generated and real target graphs. The curves of the generated graphs indeed  follow power-law correctly, and become closer to the real graphs when graph size increases, which is consistent with Table \ref{table:node degree}. This again demonstrate that our end-to-end graph translator indeed can accurately discover the underlying translation by scale-free model and its parameters. Due to space limits, many other examples can be found in Appendix A in supplementary materials.

\paragraph{Indirect Evaluation}
Table \ref{table:scale free result} shows the average graph classification measurements: Precision(P), Recall(R), AUC and F1-measure, for graph size 50, 100 and 150. ``Optimal'' denotes the classifier trained by real target graph as positive samples. It can be seen that the classifiers trained based on the targe graphs generated by GT-GAN outperforms those by randomVAE in size 50 on generating graph basically by over 10\% in all metrics. Moreover, the performance corresponding to GT-GAN is very close to the ``optimal'' in majority of the metrics.
\floatsetup[table]{capposition=top}  
\newfloatcommand{capbtabbox}{table}[][\FBwidth]
\begin{table*}
\begin{floatrow}  
\capbtabbox{  
 \scriptsize{
  \begin{tabular}{p{0.4cm}p{1.4cm}p{0.4cm}p{0.4cm}p{0.4cm}p{0.4cm}}
    \toprule
    Size& Method   & P    &R    &AUC &F1 \\
    \midrule   
    50 & randomVAE   &0.89 &0.67 &0.84   &0.76    \\
    50 & GT-GAN  &0.93 &0.82 &0.94  &0.87   \\
    50 & optimal &0.94 &0.90 &0.97 &0.91 \\
    \midrule    
    100 & GT-GAN  &0.72 &0.69 &0.68  & 0.70 \\
    100 & optimal &0.99 &0.61 &0.81 & 0.75\\
    \midrule    
    150 & GT-GAN  &0.94 &0.79 &0.96  &0.86  \\
    150 & optimal &0.99 &0.93 &0.96 &0.95\\
    \bottomrule
  \end{tabular}
  }
  }{
  \caption{Evaluation for scale-free graphs}
  \label{table:scale free result}
}  
\capbtabbox{   
 \scriptsize{
  \begin{tabular}{p{0.4cm}p{1.4cm}p{0.4cm}p{0.4cm}p{0.4cm}p{0.4cm}}
    \toprule
    Size& Method   & P    &R  &AUC &F1 \\
    \midrule
    50 & randomVAE   &0.93 &0.46 &1.00  &0.63     \\
    50 & GT-GAN  &1.00 &0.99 &1.00 &0.99    \\
    50 & optimal &0.99 &1.00 &1.00& 0.99 \\
    \midrule
    100 & GT-GAN  &0.90 &1.00 &1.00 &0.94   \\
    100 & optimal &1.00 &1.00 &1.00 &1.00\\
    \midrule   
    150 & GT-GAN  &0.97 &1.00 &1.00  &0.98  \\
    150 & optimal &1.00 &0.99 &1.00 &0.99\\
    \bottomrule
  \end{tabular}
  }
}{  
 \caption{Evaluation for Poisson random graphs}
  \label{table:poission result}
}  
\end{floatrow}
\end{table*}  
\subsubsection{Performance on Poisson Random Graph Set}
\begin{table}[h]\scriptsize
  \centering
   \caption{\vspace{-0.3cm}Classification performance for two methods in user authorization dataset}
  \begin{tabular}{llllll}                   \\
    \toprule
     Graph size & Method   & Precision    &Recall    &AUC &F1  \\
    \midrule 
    50 &randomVAE &0.32 & 0.51& 0.26 &0.39\\
    50 &GT-GAN &0.79 &0.68 &0.78 &0.73\\
    50 &optimal &0.97 &0.97 &0.97 &0.97\\
    \hline    
    300 &GT-GAN  &0.84 &0.66 &0.79 &0.74    \\
    300 &optimal &0.98 &0.96 &0.97 &0.97 \\
    \bottomrule
  \end{tabular}

  \label{table:user authorization}
\end{table}
For direct evaluation, the mean of edge increasing ratio $k$ for generated graphs by our GT-GAN is 3.6, compared to the real value of 5, which implies that the GT-GAN generally is able to discover the underlying increasing ratio between input and target graphs. More evaluation results can be found in Appendix B. For indirect evaluation, the precision, recall, AUC, and F1-measure of three classifiers are listed in the Table \ref{table:poission result}. The GT-GAN exhibit very outstanding performance, not only outperforms randomVAE by around 50\% in F1-measure, but also is highly close to the ``optimal''. This indicate that GT-GAN can effectively generate samples for classifiers which only have not any training data under some specific classes.

\subsubsection{Performance on User Authorization Graph Set}
\paragraph{Indirect evaluation}
According to Section \ref{sec:datasets}, each user needs a classifier to identify whether the specific user's behavior is normal or hacked. GT-GAN is trained by the samples from half of all the users and then generate positive samples for the other users to train their classifiers. As shown in Table \ref{table:user authorization}, classifier trained by generated graphs by GT-GAN can effectively classify normal and hacked behavior with AUC above 0.78, largely above the 0.5 if using random model. GT-GAN significantly outperforms randomVAE by around 0.9 over randomVAE on all the metrics. And its performance is consistently excellent when the graph size varies from 50 to 300.  More directed evaluation results can be found in Appendix C, including the case studies.


\section{Conclusion and Future Works}

This paper focuses on a new problem on deep graph translation. To achieve this, we propose a novel Graph-Translation-Generative Adversarial Networks (GT-GAN) which will generate a graph translator from input to target graphs. To learn both the global and local graph translation mapping for directed weighted graph, new graph convolutions and deconvolutions have been proposed to encode and decode the graphs while preserving the high-level and local-level graph patterns. Extensive experiments have been done on synthetic and real-world dataset with the comparison with the state-of-the-art method in graph generation. Experiments show that our GT-GAN indeed can discover the ground-truth translation rules, and significantly outperform the comparison method in both effectiveness and scalability. This paper opens a new window for deep graph translation. There are many possible future directions yet to explore, such as develop domain-specific graph translator using domain knowledge, or analyze and visualize the deep translation patterns.

\bibliographystyle{plain}
\bibliography{DGT}

\newpage
\appendix
\section{More experimental results for Scale Free Graph Set}
\begin{figure}[h]
  \centering
  \includegraphics[width=13.5cm] {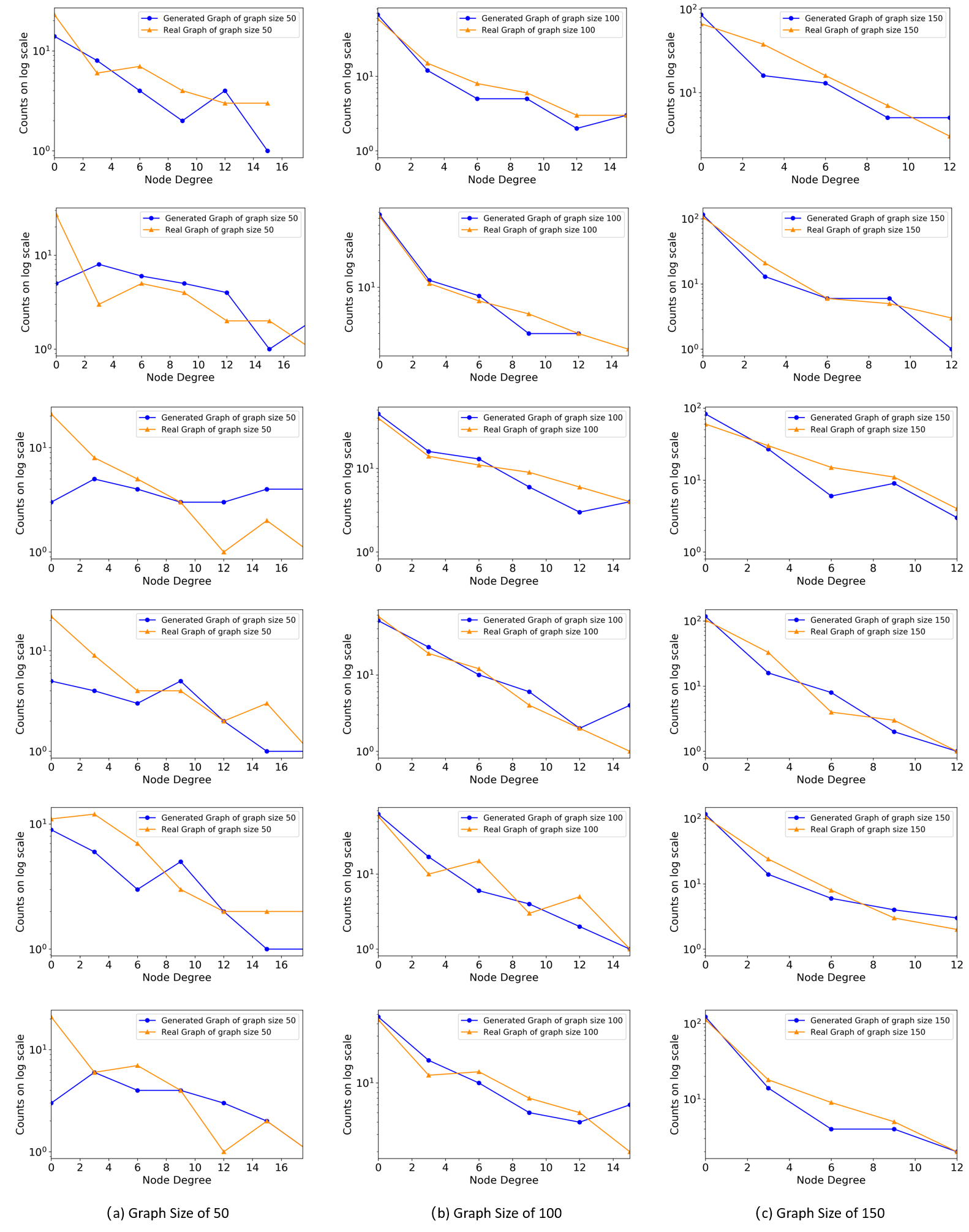}
  \caption{Examples of node degree distrbution for generated graphs and real graphs}
  \label{figure:node degree}
\end{figure}

Figure \ref{figure:node degree} shows 18 examples for scale free dataset from size 50 to 150.
\label{Appendix A}

\section{More experimental results for Poisson Random Graph Set}
\label{Appendix B}
\begin{figure}[h]
  \centering
  \includegraphics[width=10cm] {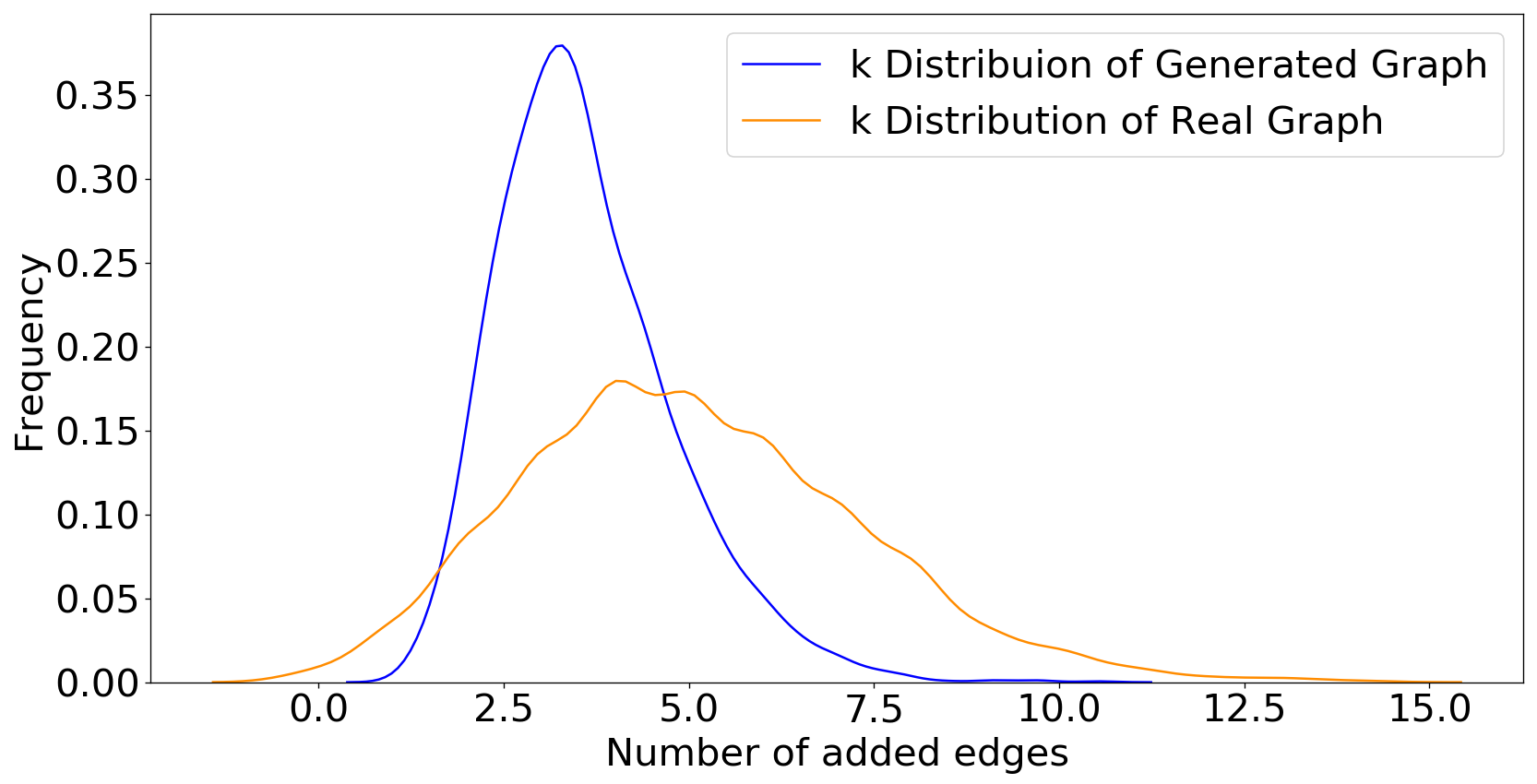}
  \caption{Distribution of $k$ for generated graphs and real graphs in Poisson random graph set}
  \label{poisson}
\end{figure}

For the Poisson random graphs, we draw the probability density curve of the proportion k. Figure \ref{poisson} shows the distribution of the $k$ in graphs generated by GT-GAN and the real graphs. The distribution plot is drew based on 3000 samples. Both of the two distribution have main degree values in the range from 2 to 7, while there is difference in the max frequency due to the limit of the samples amount. However, it prove that the proposed GT-GAN do learn the distribution type of translation parameter $k$ in this task.

Table \ref{table:poisson_direct} shows the distance measurement between generated graphs and real graphs in several metrics. For the metric "degree", we use Wasserstein distances to measure the distance of two degree distribution. For other metrics, we calculate the MSE between generated graphs and real graphs.
\begin{table}[h]
  \caption{MSE of Graph properties measurements for user authorization dataset}
  \centering
  \begin{tabular}{lllll}                   \\
    \cmidrule(r){1-5}
    Graph size & Method  &Density    &Average Degree &Reciprocity \\
    \midrule   
    50 &RandomVAE &Inf &23.68 &0.5362 \\
    50 &GT-GAN &0.0155&3.296 &0.0047\\
    \hline   
    150 &GT-GAN & 0.0142&4.373 & 0.0043 \\
    \hline   
    150 &GT-GAN & 0.0061&5.041 & 0.0019 \\
    \bottomrule
  \end{tabular}
  \label{table:poisson_direct}
\end{table}

\section{More experimental results for User Authorization Graph Set}
\label{Appendix C}
\paragraph{About Original Dataset}
This data set spans one calendar year of contiguous activity spanning 2012 and 2013. It originated from 33.9 billion raw event logs (1.4 terabytes compressed) collected across the LANL enterprise network of approximately 24,000 computers. Here we consider two sub dataset. First is the user log-on activity set. This data represents authentication events collected from individual Windows-based desktop computers, servers, and Active Directory servers. Another dataset presents specific events taken from the authentication data that present known red team compromise events, as we call malicious event. The red team data can used as ground truth of bad behavior which is different from normal user. Each graph can represent the log-on activity of one user in a time window. The event graphs are defined like this: The node refers to the computers that are available to a user and the edge represents the log-on activity from one computer to another computer of the user.
\paragraph{Direct evaluation of User Authorization Graph Set}
We use seven metrics to evaluate the similarity of generated graphs and real graphs. The MSE value are calculated to measure the similarity between two graphs in term of different metrics. \ref{table:user authorization_direct} shows the mean square error of the generated graphs and real graphs for all users.
\begin{table}
  \caption{MSE of Graph properties measurements for user authorization dataset}
  \centering
  \begin{tabular}{lllll}                   \\
    \cmidrule(r){1-5}
    Graph size & Method  &Density    &Average Degree &Reciprocity \\
    \midrule   
    50 &RandomVAE &0.0005 &0.0000 & 6.4064\\
    50 &GT-GAN &0.0003&0.0000 &0.0002\\
    \hline   
    300 &GT-GAN & 0.0004&0.0000 & 0.0006\\
    \bottomrule
  \end{tabular}
  \label{table:user authorization_direct}
\end{table}
\paragraph{Case Studies on the generated target graphs}
\begin{figure}
  \centering
  \includegraphics[width=0.9\textwidth]{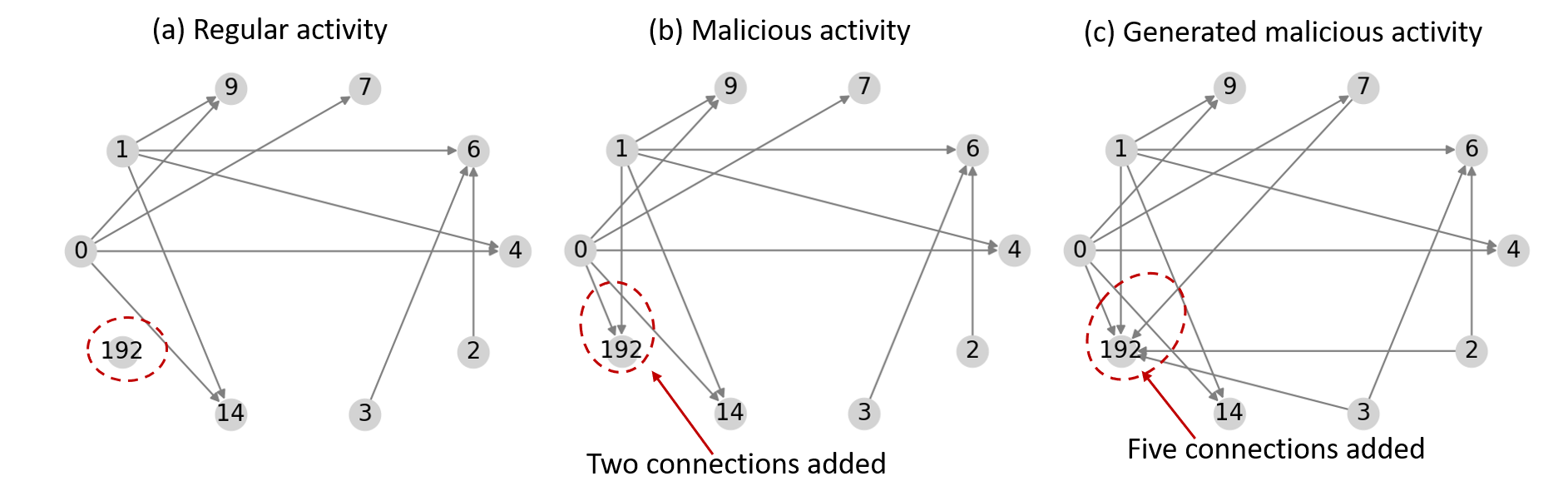}\vspace{-0.1cm}
  \caption{Regular graphs, malicious graphs and generated graphs of User 049\vspace{-0.4cm}}
  \label{case2_1}\vspace{-0.4cm}
\end{figure}\vspace{-0.4cm}
Figure \ref{case2_1} shows the example of User 049 with regular activity graph, real malicious activity graph and malicious activity graph generated by our GT-GAN from left to right. Only those of edges with difference among them are drawn for legibility. It can be seen that, the hacker performed attacks on Computer 192, which has been successfully simulated by our GT-GAN. In addition, GT-GAN also correctly identified that the Computer 192 is the end node (i.e., with only incoming edges) in this attack. This is because GT-GAN can learn both the global hacking patterns (i.e., graph density, modularity) but also can learn local properties for specific nodes (i.e., computers). GT-GAN even successfully predicted that the hacker connect from Computers 0 and 1, with Computers 7 and 14 as false alarms.

For User006, the red team attackers make more connections on Node 36 compared to user's regular activity, as marked in red rectangle. GT-GAN leans how to choose the Node 36 and it generated more connections too in the Node 36
\begin{figure}[h]
  \centering
  \includegraphics[width=13.5cm] {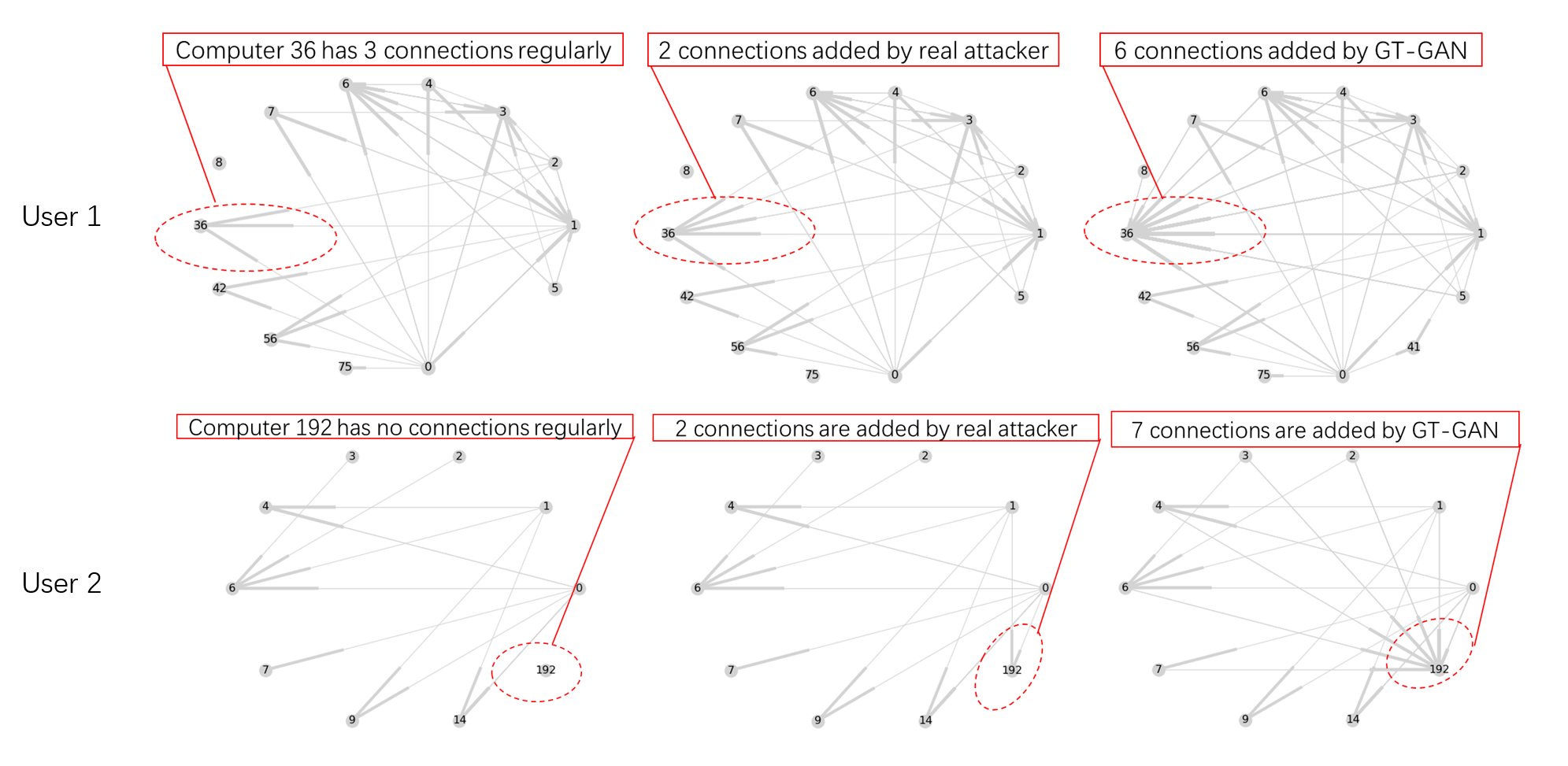}
  \caption{Regular graphs, malicious graphs and generated graphs for User 006}
  \label{case1}
\end{figure}
. 
\section{Flowchart of indirect evaluation process}
\begin{figure}[h]
  \centering
  \includegraphics[width=12cm] {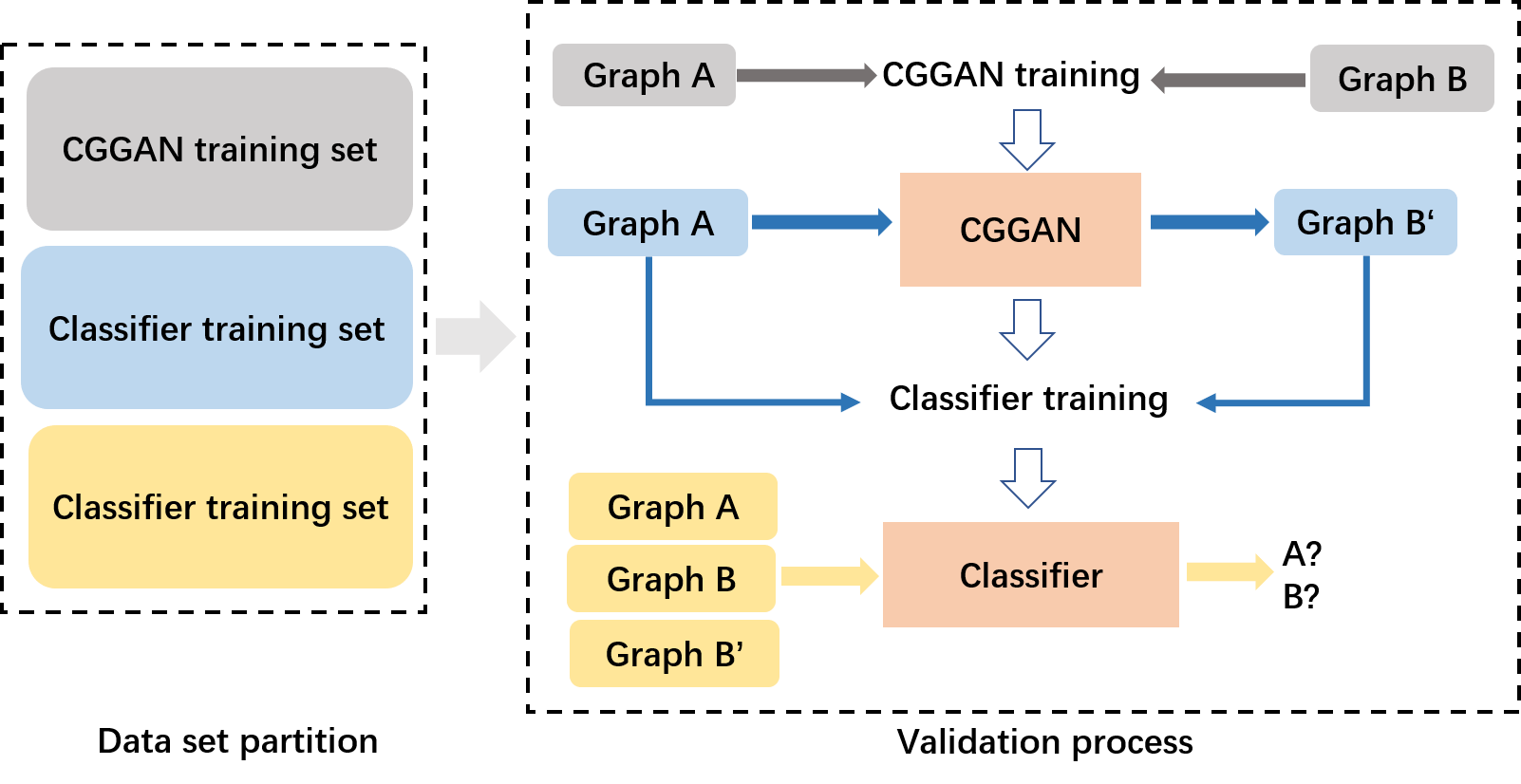}
  \caption{Flow chart of validation}
  \label{flow chart}
\end{figure}

Figure \ref{flow chart} shows the process of the indirect evaluation process.
\label{Appendix D}

\section{Architecture Parameter for GT-GAN model}
\textbf{Graph Generator}: Given the graph size (number of nodes) $N$ of a graph. The output feature map size of each layer through graph generator can be expressed as:

$N\times N\times 1 \to N\times N\times 5 \to N\times N\times 10 \to N\times 1\times\ 10 \to N\times N\times 10 \to N\times N\times 5 \to N\times N\times 1$

\textbf{Discriminator}: Given the graph size (number of nodes) $N$ of a graph. The output feature map size of each layer through graph discriminator can be expressed as:

$N\times N\times 1 \to N\times N\times 5 \to N\times N\times 10 \to N\times 1\times 10 \to 1\times 1\times 10$

For the edge to edge layers, the size of two kernels in two directions are $N\times 1$ and $1\times N$. For the node to edge layer, the kernel size is $1\times N$
\end{document}